\documentclass[journal]{IEEEtran}
\usepackage[dvips]{graphicx}
\input{epsf}
\usepackage{url}
\usepackage{subfig}
\usepackage[table]{xcolor}
\usepackage{latexsym}
\usepackage{lastpage}
\usepackage{cite}
\usepackage{color}
\usepackage{listings}
\usepackage{epsfig}
\usepackage{amsfonts}
\usepackage{amsmath}
\usepackage{amssymb}
\usepackage{mathtools}
\usepackage{algorithm}
\usepackage{algorithmic}
\usepackage{latexsym}
\newcommand{\beq}{\begin{equation}}
\newcommand{\eeq}{\end{equation}}
\newcommand{\beqr}{\begin{equation}\begin{array}{l}}
\newcommand{\eeqr}{\end{array}\end{equation}}
\newcommand{\beqa}{\begin{eqnarray}}
\newcommand{\eeqa}{\end{eqnarray}}

\newcommand{\be}{\begin{equation}}
\newcommand{\ee}{\end{equation}}
\newcommand{\bea}{\begin{eqnarray}}
\newcommand{\eea}{\end{eqnarray}}
\newcommand{\ba}{\begin{array}}
	\newcommand{\ea}{\end{array}}
\newcommand{\beas}{\begin{eqnarray*}}
	\newcommand{\eeas}{\end{eqnarray*}}
\newcommand{\leftm}{\left[\begin{array}}
	\newcommand{\rightm}{\end{array}\right]}

\definecolor{gray}{rgb}{0.4,0.4,0.4}
\definecolor{darkblue}{rgb}{0.0,0.0,0.6}
\definecolor{cyan}{rgb}{0.0,0.6,0.6}
\definecolor{applegreen}{rgb}{0.55, 0.71, 0.0}
\lstdefinelanguage{XML}
{
	morestring=[b]",
	morestring=[s]{>}{<},
	morecomment=[s]{<?}{?>},
	stringstyle=\color{black},
	identifierstyle=\color{darkblue},
	keywordstyle=\color{cyan},
	morekeywords={xmlns,version,type}
}

\begin{document}
\graphicspath{{./Figures/}} 
\title{Toward an Expressive Bipedal Robot: Variable Gait Synthesis and Validation in a Planar Model}
\author{Umer~Huzaifa,~\IEEEmembership{Student Member,~IEEE}, Catherine Maguire, and~Amy~LaViers,~\IEEEmembership{Member,~IEEE}%
		\thanks{Corresponding author, Umer Huzaifa, and Amy LaViers are with Department of Mechanical Science and Engineering, University of Illinois at Urbana-Champaign, IL 61801, USA. (email: \{mhuzaif2, alaviers\}@illinois.edu)}  \thanks{Catherine Maguire is affiliated with Laban/Bartenieff Institute of Movement Studies (LIMS), Brooklyn, NY 11217, USA. (email: catmaguire@embarqmail.com)}}      
	\maketitle
	\begin{abstract}		  
Humans are efficient, yet expressive in their motion.  Human walking behaviors can be used to walk across a great variety of surfaces without falling and to communicate internal state to other humans through variable gait styles. This provides inspiration for creating similarly expressive bipedal robots. To this end, a framework is presented for stylistic gait generation in a compass-like under-actuated planar biped model. The gait design is done using model-based trajectory optimization with variable constraints. For a finite range of optimization parameters, a large set of 360 gaits can be generated for this model. In particular, step length and cost function are varied to produce distinct cyclic walking gaits. From these resulting gaits, 6 gaits are identified and labeled, using embodied movement analysis, with stylistic verbs that correlate with human activity, e.g., ``lope'' and ``saunter''. These labels have been validated by conducting user studies in Amazon Mechanical Turk and thus demonstrate that visually distinguishable, meaningful gaits are generated using this framework. This lays groundwork for creating a bipedal humanoid with variable socially competent movement profiles.
	\end{abstract}	
	\IEEEpeerreviewmaketitle	
    
    \vspace{-.2in}
	\section{Introduction}		
Humans are capable of generating a wide range of movement behaviors, making them both efficient and expressive in walking.  Moreover, this behavior is leveraged in social interactions: by modulating gait from, for example, sluggish to peppy, we share internal state with human counterparts, saying either, for example, ``I'm tired today'' or ``I'm ready to work''. Therefore, one goal in the design of a bipedal robot may be to imitate the functional aspect of human gait, but for a social bipedal robot, it is also important to enrich the movement styles to make these robots more expressive.

Expressive robotic systems have been shown to be important in human robot interaction \cite{breazeal2000sociable,andrea2016soro,hamacher2016roman,Leite2014,Rossi2018}; for example, movable facial features enabled the robot Kismet to interact with human counterparts. One approach to creating expressive bipedal walkers has been to add faces to existing humanoid platforms \cite{kuroki2003small,oh2006design,anaso2011waci,kishi2012development,kishi2013development,Liu2013,asheber2016ijrr,Stephens-Fripp2017}. This augmentation uses facial expression, similar to Kismet, to indicate internal state of the system.  This technique does not modify the motion of the walking gait toward social goals, but this behavior is part of how humans behave in social settings \cite{Cutting1977,davis2001visual,blake2007perception,putten2018}.  In particular, body motion augments facial expression in human behavior \cite{van2007body}.
       
The motion of both passive and active bipedal robots have largely been designed in a functional vein. One of the seminal bipedal robots built \cite{mcgeer} was a simple, passive structure for efficient walking. In this work, an analysis  was presented on the design parameters for a passive walker with stable gait on a downward slope. Subsequently, a three dimensional passive walker version was developed in \cite{collins2001three}. The main goal for designing such robots was to obtain the energetically efficient movement in robot given no or minimal energy input. 
	
The category of actively controlled bipedal robots can be further subdivided. Categories include statically stable and dynamically stable robots. A comparison of these two types of active walking robots has been presented in \cite{AKuo_compare}. Among the statically stable robots are mostly humanoid robots which can be made stable by maintaining a posture that satisfies static equilibrium properties. More specifically, they are designed to keep their so called zero moment point (ZMP) inside the polygon made by their feet \cite{ZMP_35yrs}. Robots using this type of control include ASIMO, ATLAS and NAO \cite{ASIMO,feng2014optimization,G2009}.
	
Dynamically stable robots, on the other hand, have to be in motion to keep themselves from falling. Examples for these include \cite{ranger} that achieved passivity-based walking with a minimal input, and walked on flat ground for a record distance on a single battery charge. Other examples inspired from passive walking, executing dynamically stable motion, include \cite{Spong02controlledsymmetries}, \cite{Denise}, \cite{Hobbelen2008a} and \cite{goswami}. A number of robots in this category use  linearization based control schemes to make the robot follow a limit cycle behavior. These robots include MARLO \cite{marlo}, MABEL \cite{mabel}, AMBER \cite{amber}, among others. In these methods, robotic legs are extended by closely following a desired reference defined in terms of a state variable. Using this framework, in \cite{Hereid_ICRA16}, trajectory optimization is used to compute stable gaits for walking. A related application is in the lower limb exoskeletons where optimized joint reference trajectories are followed by the joints \cite{Huang2016}.

There has been considerable work done on developing variety for functional purposes in walking gaits on a bipedal walking robot as well. Among the latest work is \cite{da20162d} where a precomputed gait library for different gait speeds is prepared and after developing a map using learning techniques, these gaits are employed at runtime on a 3D bipedal robot. In another recent work, \cite{motahar2016composing}, gait primitives in terms of the stable limit cycles have been defined and their composition is studied to achieve navigation through a cluttered environment. 

In an attempt of generating different styles of walking, a catwalk was investigated on an HRP-2 robot \cite{catwalk_HRP2}.
Other related examples include \cite{ParkJH_ICRA2003} in which different ways of locomotion of bipedal robot and transitions between them were investigated. In \cite{Atkeson_RAM2007}, under variations in an observed walking behavior, different stable walking gaits were developed using reinforcement learning on a biped robot. In the community of computer animations, SIMBICON \cite{Yin07} is another control scheme for generating different bipedal walking behaviors. 

The link between affect and emotion in human gait has been explored via a computational model built to  recognize different emotions in human gaits in \cite{venture2014recognizing}. In \cite{etemad2016expert}, a framework was developed using motion capture data to generate different walking styles associated with different emotions with the help of animation experts, followed by validation in a user study. Yet another example is \cite{Kavafoglu2016} in which using high-level features identified from motion capture data, different gait styles are generated in a physics-based simulation which are validated as well.  In \cite{asanoSMC2001}, a walking pattern generator is proposed that gives variable walking motions in a planar biped model in response to different disturbances. Further work has shown that in uncertain scenarios, finding variation in movement can be viewed as an optimal strategy \cite{todorov2002optimal}. 

One place the social aspect of walking is seen is in crowds.  Humans take signals from the environment, as well as the manner of walking of those around them.  In an emergency setting, humans notice their counterparts' quickened pace and tensed muscles and respond, changing their own gait in kind.  This has been studied as an important example of social behavior affecting public safety \cite{gallup2012visual}.  Further, walking robots in any human-facing environment will likely be attributed social attributes regardless of whether their gaits have been intentionally shaped \cite{hortensius_cross_2018}.  Parallel work has shown that perception of walking various across environmental contexts but that this effect can be mitigated through variable styles of walking \cite{heimerdinger2017influence}.

Thus, we are presenting a framework that allows generating a wide variety of walking movements in a simple planar biped model for expressing the internal state and communicating with human counterparts. Specifically, this paper investigates the expressivity of a set of generated gaits, by labeling them with suitable verbs in English associated with different human walking styles. These label are then verified for corresponding gaits by lay viewers through a user study. 

The rest of the paper has the following structure. First, we discuss studying variation in walking motion through the lens of embodied movement analysis and using language as a tool to convey it. In Section \ref{sec_model}, we go over the mathematical modeling of the under-actuated planar biped model used for this work. For gait synthesis in this model, a model-based trajectory optimization method is introduced in Section \ref{sec_control}. In Section \ref{sec_synthesis}, the process of selecting a set of gaits for validation and labeling them is discussed. Section \ref{sec_validate} discusses the results of validation and details of the user study.  Finally, Section \ref{conc} discusses how this work contributes toward the development of a social, expressive bipedal platform.

\begin{figure*}
	\centering
	\includegraphics[width=2\columnwidth]{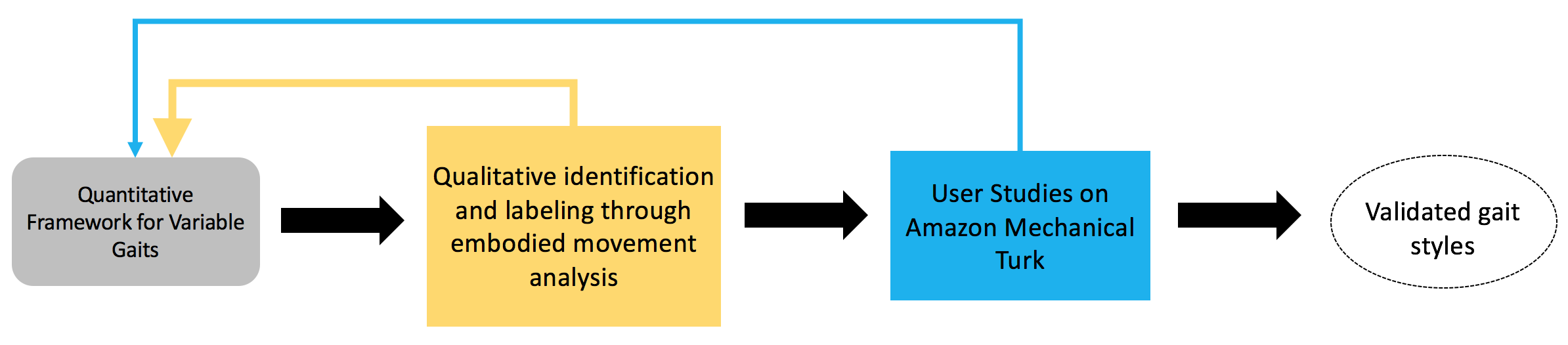}
    \caption{Process of how the gait styles along with their labels have been identified.  The role of human observation will be leveraged in two ways: first, expert movement analysts work iteratively to inspect and come to bodily understanding of the produced gaits (Section \ref{sec_synthesis}); then, lay viewers confirm, or challenge, this analysis (Section \ref{sec_validate}).  This two part process creates two iterative design cycles in our workflow.}
    \label{fig:style_process}
\end{figure*}

\section{Style Analysis Through a Choreographic, Embodied Lens}
		\label{fundamentals}	        
We use the embodied movement taxonomy in the Laban/Bartenieff Movement System (LBMS) to understand walking and frame the goals of this work. This taxonomy allows us to investigate, from a bodily perspective, all the ways we might vary our gait in order to accommodate environment and/or task. That means, as a research team, trying out various gaits on ourselves, moving across the room from various movement prompt, e.g. ``skitter''. Then, we look to literature and language references to understand how various forms of walking are described by and communicated to humans. This work forms the basis of our labeling of variable gaits.

Walking is a specific instance of Locomotion, one of the Basic Body Actions enumerated in LBMS, that is characteristic to human movement behavior \cite{studd2013}. Simply put, it is the mobilization of our weight through space in order to change location through bipedal action that leaves at least one leg on the ground at all times. In order to move our weight, we must shift our center of weight over, to and between our two legs. In walking this is an ongoing dynamic that forms a stereotyped action of patterning of the bipedal form where one foot is always in contact with the floor. 

To this end, there are a wide variety of gaits that human movers (and robots) can produce based on form, intent, context and phrasing. This variety accommodates walking inside of different kinds of environments \cite{Bartenieff}: slippery ice, rocky terrain, and even say over hot coals as well as walking motivated by different internal states: a particular mood, intent, or motivation. Thus, ``walking'' can be further articulated into multiple identifiable gait styles.

In practice, this action is based on environmental constraints and communication with human counterparts concerns as well.
Take for example a person in high heels negotiating a crowded sidewalk, versus a person in sneakers. The context of the clothing will affect the gait. The stride length, the heel strike, and the transfer of weight will manifest differently based on the footwear.  The intent of the mover will also change the gait. For example, dragging one’s feet to avoid arriving at an undesired confrontation or 
tiptoeing over a noisy floor to avoid making sounds,  will change the style of walking.

In addition to being identified inside movement theory, these styles show up in language. The use of these different gait words evokes particular images and suggests an attitude, experience and intent of the mover. It also portrays a particular relationship to the environment, or context of the mover. There are a multitude of examples of this both in the spoken and written language. Nicholson writes of ``many synonyms, or not quite synonyms, for walking, each word with its own shade and delineation of meaning'', going on to discuss his own relationship to as many as 23 different words for walking, e.g., ``tromped'', ``strolled'', and ``hiked'', and then carrying on to discuss prepositional modifiers too, e.g., ``walked on eggshells'' \cite{nicholson2008lost}.  Merriam-Webster’s thesaurus lists 62 words that are related to or synonyms of walking \cite{mw}.

The resulting gaits can be identified through phrasing and are recognizably separate from each other. These walking styles are interpreted inside of context by human viewers in order to estimate the internal state of the mover. Our interest is in seeing if we can produce different, and recognizable, gaits in a bipedal robotic platform. Identifying, naming, and validating the gaits (Fig. \ref{fig:style_process}) is a tool toward this end.

\section{Planar Biped Model}
\label{sec_model}
In this section, a planar model of an under-actuated bipedal robot is presented. The structure of this model is inspired from the sagittal plane (side) view of a human walking. In this model, ``legs'', without ``knees'', are considered.  The legs can rotate about the ``hip'', with point masses at their centers of mass. This model is commonly known as compass-like biped model and has been studied extensively  in the literature, e.g., \cite{goswami1996compass}. 

In the traditional version, this model is passive, i.e., without any actuation, walking on an incline. In comparison, we are considering it on a flat surface with actuation in one of the legs. Such a leg configuration without knees on a flat surface causes leg scuffing but it is assumed, as in \cite{grizzle1999poincare}, that through some external actuation, the swing leg moves in the coronal plane and returns to the sagittal plane at the time of impact only. Furthermore, it is assumed that the swing leg does not experience slip or rebound upon impact with the ground.

Using the guidelines presented in \cite{grizzle_book}, the biped model is defined for the swing phase (when one of the legs is off-ground) and the strike phase (when both legs touch the ground) as follows in the next subsections.

\begin{figure}[h]
	\centering
	\includegraphics[width=\columnwidth]{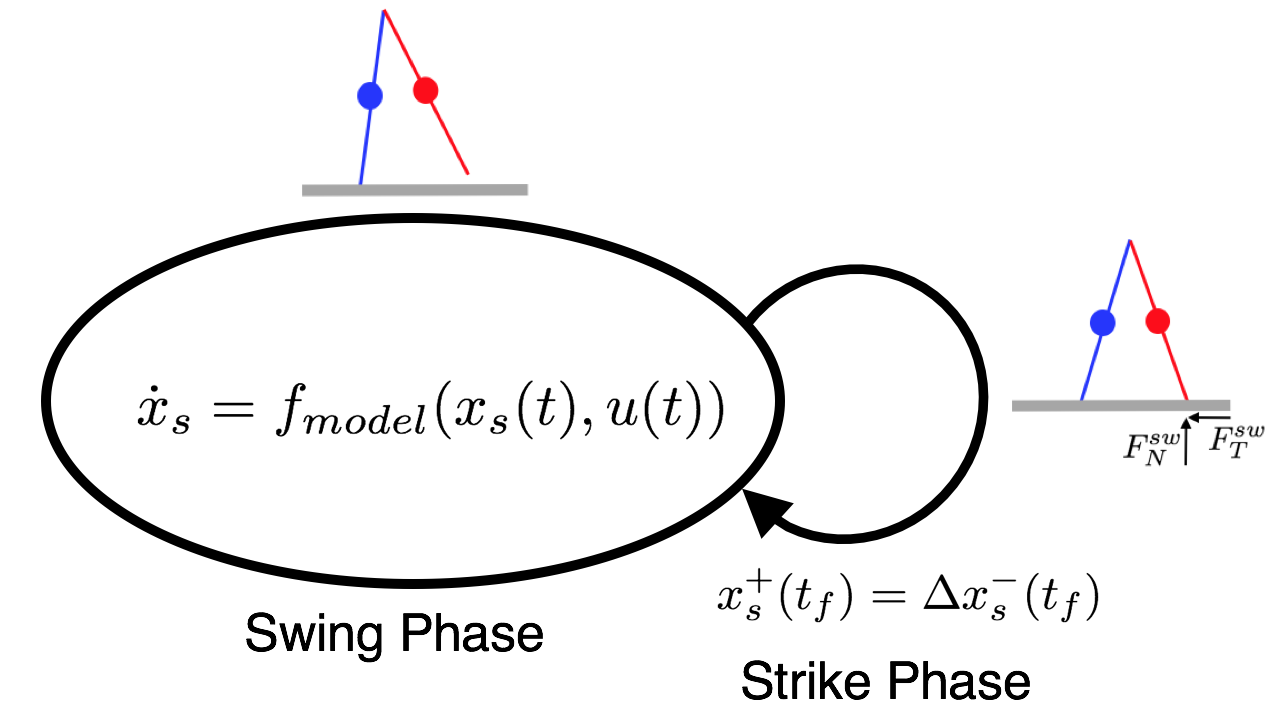}
	\caption{The stable walking of the planar biped model can be represented as a hybrid system. The biped model stays in the swing phase dynamics until the swing leg impacts the ground. $F^{sw}_N$, and $F^{sw}_T$ are the ground reaction forces acting at the swing leg foot. At that instant, the biped state variables are updated using strike phase dynamics and returned to the swing phase.  As a result, the right and left legs swap their roles.}       
    \label{hybrid} 
    \vspace{-0.2in}
\end{figure}

\paragraph{Swing Phase Dynamics}

In the swing phase, the biped model acts as a two link planar robot fixed at the stance leg foot. The Euler-Lagrange equation for this phase is as follows:
\begin{align}\label{eul_lag}
 \frac{d}{dt} \frac{\partial{\mathcal{L}_s}}  {\partial{\dot{q_s}}}-\frac{\partial{\mathcal{L}_s}}{\partial{q_s}} = \Gamma_s,
\end{align}
As a result, we obtain the equations of motion:
\begin{align}\label{swing_model}
D_s\ddot{q_s}+C_s(q_s,\dot{q_s}) \dot q_s+G_s=\Gamma_s 
\end{align}
where $q_s=[q_{st},q_{sw}]^T$ is the set of generalized coordinates, \textit{$D_s$} is the inertial matrix, \textit{$C_s$} represents the Coriolis and Centrifugal terms, and \textit{$G_s$} are the gravitational generalized forces. These matrices are given in Appendix \ref{matrices}. The variable $\Gamma_s$ represents the generalized forces acting on the robot:
\begin{flalign}\label{gamma_s}
	\begin{split}
          \Gamma_s &= 
			\begin{bmatrix}
				0 \\ 
				1 
            \end{bmatrix}			
				\tau                    	
    \end{split}
\end{flalign}    					
where $\tau$ is the input torque in the swing leg. The state space representation of Eq. \ref{swing_model} above can be written as follows:
\begin{align}
	\begin{split}
		\dot{x}_s:=f_{model}(x_s(t), u(t))=
	      \begin{bmatrix}
			\dot q_s \\
			\ddot{q_s}
		  \end{bmatrix} \\
		= D_s^{-1}(-C_s\dot{q_s}-G_s+\Gamma_s)
    \end{split}
\end{align}
\paragraph{Strike Phase Dynamics}

The swing model is active until the swing leg impacts the ground. Resultantly, an update occurs in the joint positions and joint velocities. An impact model is thus obtained for the robot using Euler-Lagrange equation, whereby states right after the impact can be related to the states before the impact. We can use the following model when both of the legs are touching the ground:
\begin{align} \label{impact_model}	D_e(q_e)\ddot{q_e}+C_e(q_e,\dot{q_e})\dot{q_e}+G_e(q_e)=\Gamma_e,
\end{align}
where $q_e = [q_{st},q_{sw},p^{st}_{x},p^{st}_{y}]^T$ is the state variable, \textit{$D_e$}, \textit{$C_e$} and \textit{$G_e$} are the matrices for inertial, Coriolis and Centrifugal, and the gravity effect terms, respectively. These matrices are also provided in Appendix \ref{matrices}. For modeling impact dynamics, we need position of a point on the robot (which in this case is the stance leg foot position represented by $(p^{st}_{x},p^{st}_{y})$ to determine the reaction forces at the swing leg end. The variable $\Gamma_e$ represents the generalized forces acting on the robot. 
\begin{align}\label{impact_gamma}
	\Gamma_e = 
		\begin{bmatrix}
			\Gamma_s \\ 
			0 \\
			0
		\end{bmatrix}	
        + \delta F_{ext},				
\end{align}
where $\Gamma_s$ is given by Eq. \ref{gamma_s} and $$F_{ext} = (E(q^-_e(t_f)))^T \begin{bmatrix} F^{sw}_T \\ F^{sw}_N \end{bmatrix},$$ with $$E(q) = \frac{\partial p^{sw}(q)}{\partial q_e}.$$ Here, $\delta F_{ext}$ represents the impulse of external force applied on the biped as a result of the impact, and $p^{sw}$ represents the position of the swing leg foot. During the impact, in the terminology used by \cite{Hurmuzlu_1994}, it is assumed that the stance leg lifts off without any interaction with the ground; therefore external force on the stance leg is considered to be zero. Tangential and normal components of external force, $F^{sw}_T$ and $F^{sw}_N$, respectively, are shown in Fig. \ref{hybrid}. Assuming that the impact happens at $t=t_f$, integrating Eq. \ref{impact_model} over the interval of impact, we obtain:\\
\begin{equation}
      \begin{split}
 	    \label{constrt_1}
          D_e(q_e^-(t_f))(\dot q^+_e(t_f) - \dot q^-_e(t_f))=F_{ext} \\ = (E^T(q^-_e(t_f)))^T 
          \begin{bmatrix} 
             F^{sw}_T \\ F^{sw}_N 
          \end{bmatrix},
      \end{split}			                 \end{equation}
where it is assumed that during the impact, the joint positions remain continuous. For stable walking, we need the swing leg to come to rest at the end of a step, keeping its impact position. This condition can be given as follows: 
\begin{align} \label{constrt_2}
	E(q^-_e(t_f)) \dot q^+_e(t_f) = 0.
\end{align}				
At this moment, the joint positions undergo relabeling and are updated as follows:
\begin{equation}
	\begin{split}		
      q_{st}^{+}(t_f) = q_{sw}^{-}(t_f), \\
      q_{sw}^{+}(t_f) = q_{st}^{-}(t_f),     
    \end{split}
\end{equation}
where the positive sign on top of these variables represents updated values after the impact and the negative sign shows the values before impact. 

As evident from the above equations, the two leg joint angles simply change their roles. The new swing leg acquires a joint velocity given by the following expression (derived by solving Eq. \ref{constrt_1}, and \ref{constrt_2}):
\begin{equation} \label{reaction}
	\begin{split}
	  \begin{bmatrix}
		D_e(q_e^-(t_f))  &  -E(q^-_e(t_f))^T  \\
		E(q^-_e(t_f))  & 0  \\
	  \end{bmatrix}
	  \begin{bmatrix}
		\dot{q}_e^+(t_f)    \\
		F^{sw}_T    \\
		F^{sw}_N    
	  \end{bmatrix} \\
		=
	  \begin{bmatrix}
		D_e(q_e^-(t_f))\dot{q}_e^-(t_f)   \\
		0    \\
	  \end{bmatrix}
   \end{split}
\end{equation}
and can be expressed as follows for the swing leg joint states:
\begin{flalign}
	x_s^+(t_f) = \Delta (x_s^-(t_f))			\end{flalign}
where, $\Delta$ is an instantaneous mapping from the biped state just before impact with the ground i.e., $x^-(t_f)$ to the biped state right after the impact i.e., $x^+(t_f)$. As seen in Fig. \ref{hybrid}, the biped stays in the swing phase, as long as the swing leg does not impact the ground. At that instant, the biped goes through the strike phase, and undergoes update in the joint angles and velocities. The updated states are fed back to the \textit{swing phase dynamics} as the new initial states as shown in Fig \ref{hybrid}. 
    
\section{Computational Method for Stable Gait Design}  		\label{sec_control}	        
For the presented planar biped model, finding a stable walking gait is formulated as a model-based trajectory optimization problem over a single walking step. This optimization problem is defined over constraints needed for walking by the biped model and an objective function. We define two types of constraints. The first type is termed as  path constraints which is satisfied throughout the walking step. The second type is boundary constraints which has to be satisfied at the two ends of a walking step. 

Essential path constraints for walking in the given model are the following:		
\begin{itemize}
	\item Swing leg dynamics of the biped model: \\ $\dot x_s = f_{model}(x_s(t), u(t))$. 			
    \item Normal ground reaction force at the stance foot remains positive: $F^{st}_N(t) >0$.
	\item Ratio of the normal ground reaction force to the tangential ground reaction force at the stance foot satisfies: $|\frac{F^{st}_N(t)}{F^{st}_T(t)}|\leq \mu$, where $\mu$ is the coefficient of friction for the walking surface.
	\item Actuator and state variable limits of the model:\\ $u_{min}\leq u\leq u_{max}$, ${x_s}_{min}\leq x_s\leq {x_s}_{max}$.
\end{itemize}
		
Essential boundary constraints for the given model, on the other hand, include the following:
\begin{itemize}
	\item Swing foot position ($p^{sw}$) at the end of a walking step reaches the desired step length defined by $L^{des}$:\\ $p^{sw}(t_f) = \begin{bmatrix} 
    	L^{des} & 0 		
    \end{bmatrix}^T.$
	\item Periodicity constraint relating the initial state and the state after impact in a walking step:\\ $x_s^+(t_f) = \Delta(x_s^-(t_f))= x_s(0)$, \\ where $\Delta$ is the map from the state before impact to the state after impact.
\end{itemize}		

Using these constraints, the formulation of optimization of a cost function $J(u(t))$ over a walking step is as follows:
\begin{equation}
	\begin{aligned}
            \min_{u(t)} & \quad  J(u(t))\\
            \text{s.t.} & \ \ \dot{x}_s=
             D_s^{-1}(-C_s\dot{q_s}-G_s+\Gamma_s) \\
            & F_N^{st} >0, \ F_T^{st} < \mu F_N^{st}\  \textrm{and} \ F_T^{st} > -\mu F_N^{st}\\     
			& r(\sin(q_{st}(t_f)) - \sin(q_{sw}(t_f)))= L^{des}\\
            & r(\cos(q_{st}(t_f)) - \cos(q_{sw}(t_f)))= 0\\
            & x_{s}(0) = \Delta (x_s(t_f))\\
            & {x_s}_{min} \leq x_s \leq {x_s}_{max} \\
            & u_{min} \leq u \leq u_{max} \\
            & t \in [0, t_f]
            \label{num_optim}
    \end{aligned}
\end{equation}

One way to solve this optimization is by discretizing the input and state trajectories for a given time duration (walking step time in our case). In this approach, the given optimization problem is discretized at specific time instants, called collocation points into a nonlinear parameter optimization problem as in \cite{Betts2009}. 

After discretization, all constraints are evaluated at the collocation points and the system dynamics model $f_{model}(x_s(t),u(t))$ is written as a set of collocation constraints. An example collocation constraint using trapezoidal method of integration is as follows:
\begin{equation}
			x^{k+1}_s - x^k_s - \frac{h_k}{2}(f_{model}^{k+1} + f_{model}^{k}) = 0, \ k \in[0, v-1],
\end{equation} 		
where $x^{k+1}_s$ and $x^k_s$ are the biped states at collocation points $k+1$ and $k$, respectively, $f^{k+1}_{model}$ and $f^{k}_{model}$ are the corresponding values of the dynamics model $f_{model}$ at these biped states, $h_k$ is the time step between the two collocation points, and $v$ is the total number of collocation points. After collocation, the discretized state and input variables are stacked in a single vector $z$, such that: 
\begin{equation}
	z =
		\begin{bmatrix}
        	x^{(1)}_s, \hdots, x^{(v)}_s \\
           	u^{(1)}, \hdots, u^{(v)}
        \end{bmatrix}.
   \label{z_def}
\end{equation}

The limitations of a standard nonlinear optimization problem are carried over to our gait finding problem too. Therefore it is possible that for some set of constraints, the solution to the optimization problem may not exist. For handling such cases, one can use some techniques to modify the problem formulation as guided in \cite{Betts2009}.		

For the given biped model, the optimization problem has been solved using a particular type of collocation, called direct orthogonal collocation, in which the discretization is carried out at time instants where the roots of Legendre polynomials exist. The nonlinear optimization problem is then solved using IPOPT \cite{IPOPT} and the function approximation is performed using the Legendre polynomials. The optimization toolbox used for this purpose is GPOPS II \cite{patterson2014gpops} and is run on a laptop computer running a 2.2 GHz Core i7 processor. The code is written in MATLAB using the baseline code structure in \cite{optimTraj} and can find solution for our optimization problem within a minute. 

\section{Synthesis of Variable Gait Styles} 
\label{sec_synthesis}
\begin{figure}
	\centering
	\includegraphics[width=\columnwidth]{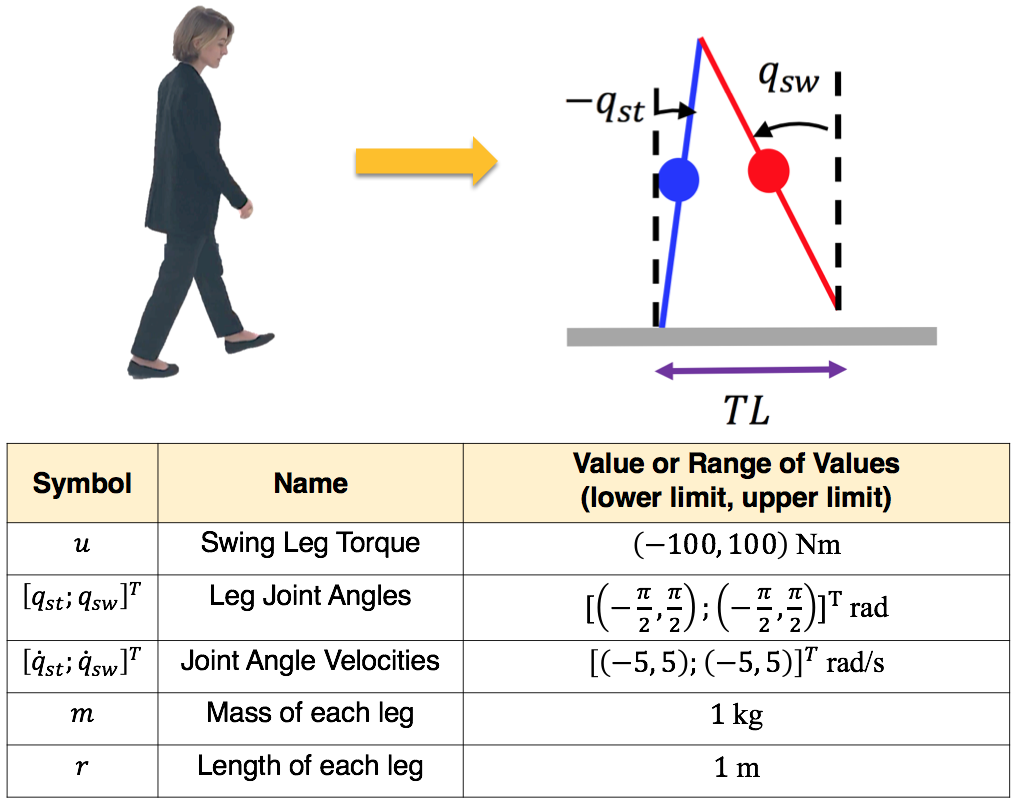}
\caption{This figure shows the analogy between human walking and the planar model (in order to justify the bodily naming of various features of the model).  The image at top was used to establish correspondence between human walking and the planar model.  The parameter $TL$ relates to Thigh Lift from the Basic Six in Bartenieff Fundamentals and assigns step length for the biped model. In the table below, range of input torque, state vector, and values of model parameters are given.}
    \label{fig:knobs+vals}
    \vspace{-0.1in}
\end{figure}
The goal for this work is to generate a wide variety of gaits, even with a low (two) degree of freedom model. This palette of gait styles can pave the way for a more expressive bipedal platform. To this end, model-based trajectory optimization is used to produce a large range of feasible gaits for suitable changes in the path constraints and the cost function. 
\begin{figure*} [t!]
	\centering
	\subfloat[]{\includegraphics[width = .41\columnwidth]{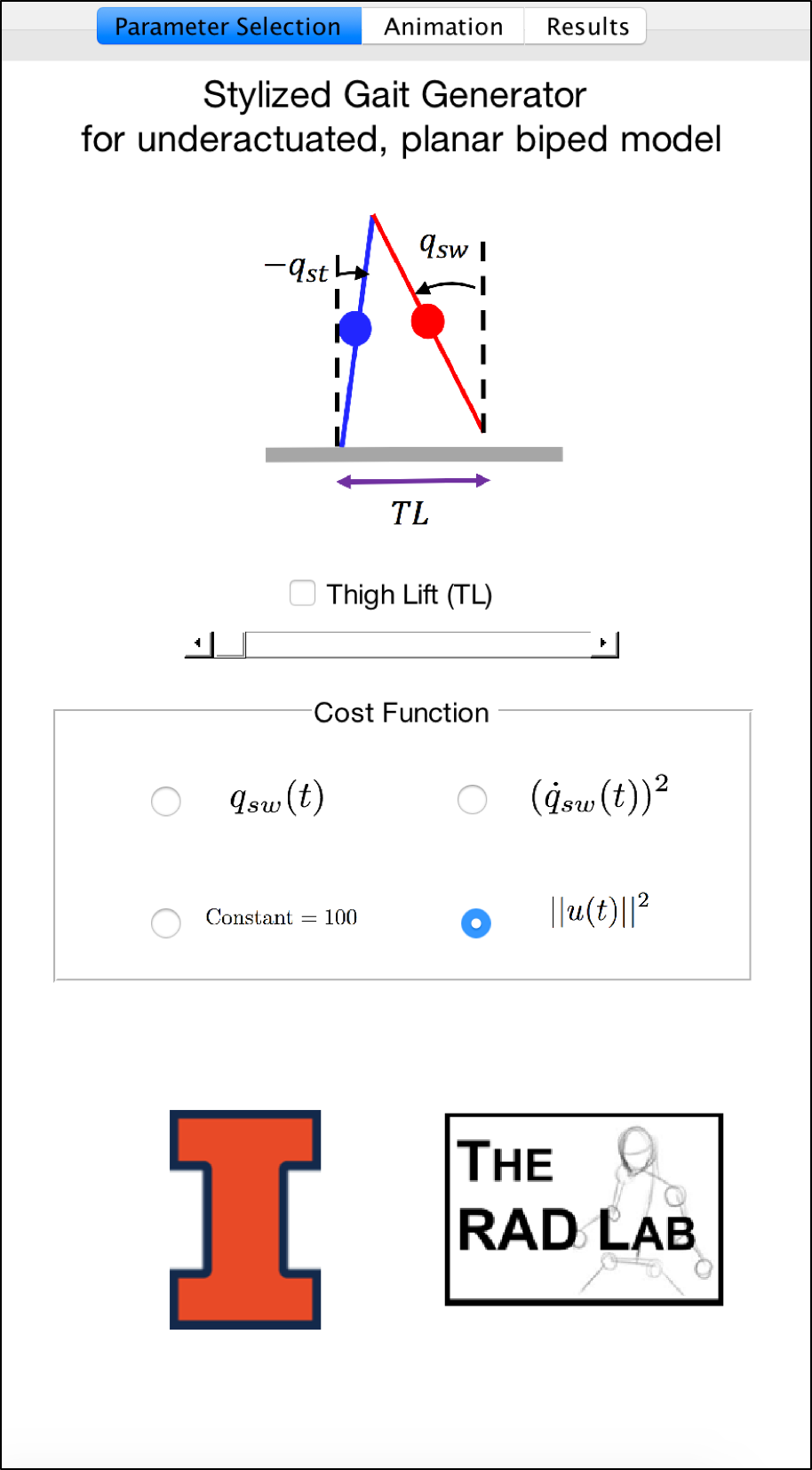}\label{fig:int_param}}
	\qquad
	\subfloat[]{\includegraphics[width = .4015\columnwidth]{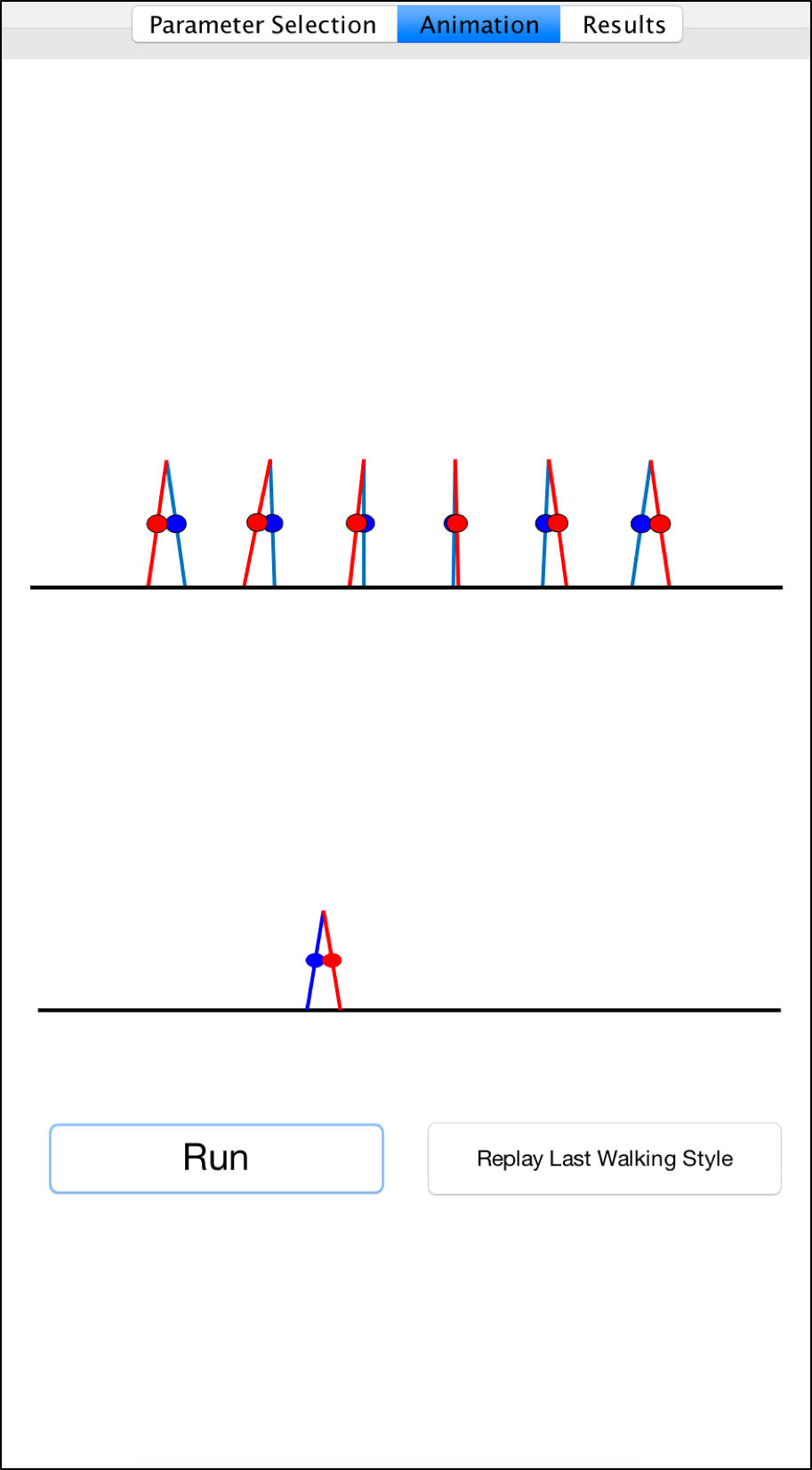}\label{fig:int_anim}}		
	\qquad
 	\subfloat[]{\includegraphics[width = .735\columnwidth]{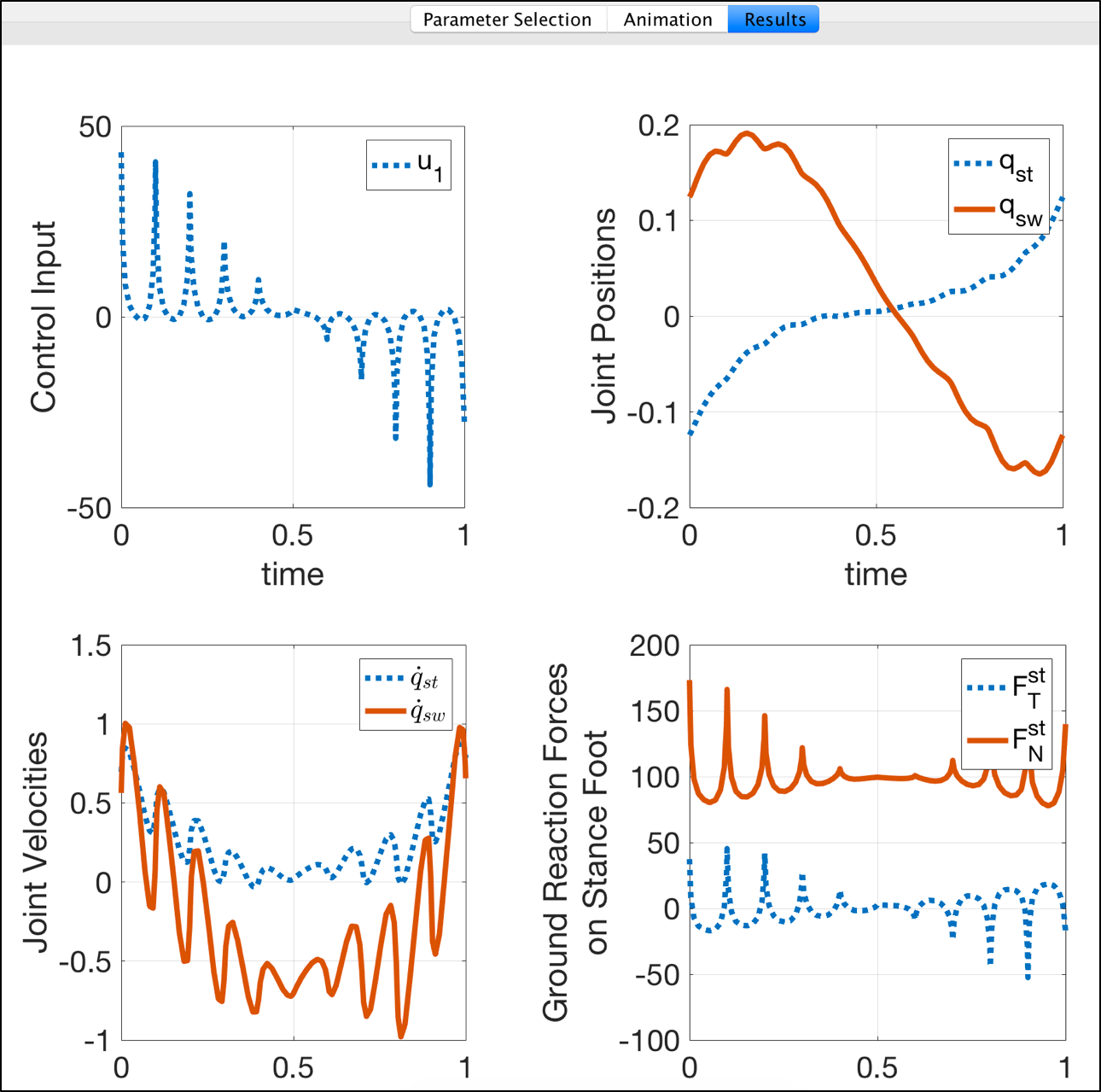}\label{fig:int_results}}    
	\caption{An interface implementing the framework given in Section \ref{sec_synthesis} that allows for iterative development of labeled gaits. The interface allows a user to define parameters in Fig. \ref{fig:int_param}. Animations of the resulting gait appears in the animation tab in Fig. \ref{fig:int_anim} and Fig. \ref{fig:int_results} shows results tab of the interface showing plots of control inputs, state variables, and reaction forces.}
	\label{fig:interface}			
\end{figure*}	

Two gait parameters, $TL$ and $cost$, are defined to generate different gaits for the given model. The optimization problem solved for finding the gaits in terms of these parameters, in bold, is defined as: \\    	 
\begin{equation}
	\begin{aligned}
			\min_z & \quad J(z) = \boldsymbol{cost} \\
            \text{s.t.} & \ \ \dot{x}_s=
             D_s^{-1}(-C_s\dot{q_s}-G_s+\Gamma_s) \\
            & F_N^{st} >0, \ F_T^{st} < \mu F_N^{st}\  \textrm{and} \ F_T^{st} > -\mu F_N^{st}\\ 
            & r(\sin(q_{st}(t_f)) - \sin(q_{sw}(t_f)))= \boldsymbol{TL}\\
            & r(\cos(q_{st}(t_f)) - \cos(q_{sw}(t_f)))= 0\\
            & x_{s}(0) = \Delta (x_s(t_f))\\
            & z_{min} \leq z \leq z_{max}
             \label{gait_set}
    \end{aligned}
\end{equation} 

\noindent where $z_{min}$ and $z_{max}$ are the limits on $z$, defined in Eq. \ref{z_def}.

The solution to this optimization problem has, empirically, been found to exist for $0.1 \leq TL \leq 0.9$, while choosing $cost$ to be one of the cost functions in $\{q_{sw}(t), (\dot{q}_{sw}(t))^2, 100, ||u(t)||^2\}$. For finding the solution, the initial and final state in the state trajectory are initialized to:
\begin{equation}         
	\begin{aligned}
    	x_s(0) = [-0.17, 0.34, 1.44, 0.53]\\ 
        x_s(t_f) = [-0.34, -0.17, 1.66, -3.25].
	\end{aligned}
\end{equation}
The biped model parameters and the range of values for state and input vectors are given in Fig. \ref{fig:knobs+vals}. As shown in this figure, the chosen value of $TL$ is used as the step length of gait and is used as  path constraint in optimization.

A user interface has been developed that allows the team to choose  parameters $TL$ and $cost$ for generating corresponding gaits and viewing the results. The interface (as shown in Fig. \ref{fig:interface}) has three tabs. In Parameter Selection tab, the parameters $TL$ and $cost$ can be selected using a slider and radio buttons, respectively. These choices, in turn, define the step length path constraint and the cost function for the optimization problem, respectively, as previously described. As a result, the optimization problem in (\ref{gait_set}) is solved and the animations are shown in Animation tab (Fig. \ref{fig:int_anim}). In Results tab, graphical results for the obtained stable gait are presented (Fig. \ref{fig:int_results}). Using this interface, we are able to generate 360 gaits for a resolution of 0.01 on the slider in Fig. \ref{fig:int_param}.

This method was used to explore different variations in biped walking motion that could be related to human walking experience. For verifying this goal, our strategy was to choose a set of gaits that could be described with suitable gait labels and then verified by a small number of lay viewers. This led to an iterative process of selecting gait labels and verifying them before we finalized a set of six labeled gaits. A schematic representation of this approach is given in Fig. \ref{fig:style_process}.

In the first step, a set of gaits was selected out of the total 360 gaits. These were selected through extensive observation of the available gaits to be visually distinguishable from each other. In the next step, verbs describing various walking styles were assigned to the selected gaits.

This assignment was carried out by observing the gaits multiple times going back and forth between a macro and a micro perspective. The macro perspective focused on the ``phrasing'' of the pattern: its rhythm, duration, and emphasis. These determinations were made by looking at the movement of the parts in relationship to each, and how the parts, depending on their phrasing, created a whole gait style. In order to further clarify and assign labels to the animations, an embodied process followed. We ``tried'' on the animations and replicated them in our own bodies to understand the associations and experiences that the perceived phrasing patterns generated. So, through extensive observation, identification of phrasing patterns and an embodied experience of the animations, a link was made to the larger body of knowledge of ``types'' of human locomotion. 

\begin{figure*} [h!]
    \centering	
    \includegraphics[width=2\columnwidth]{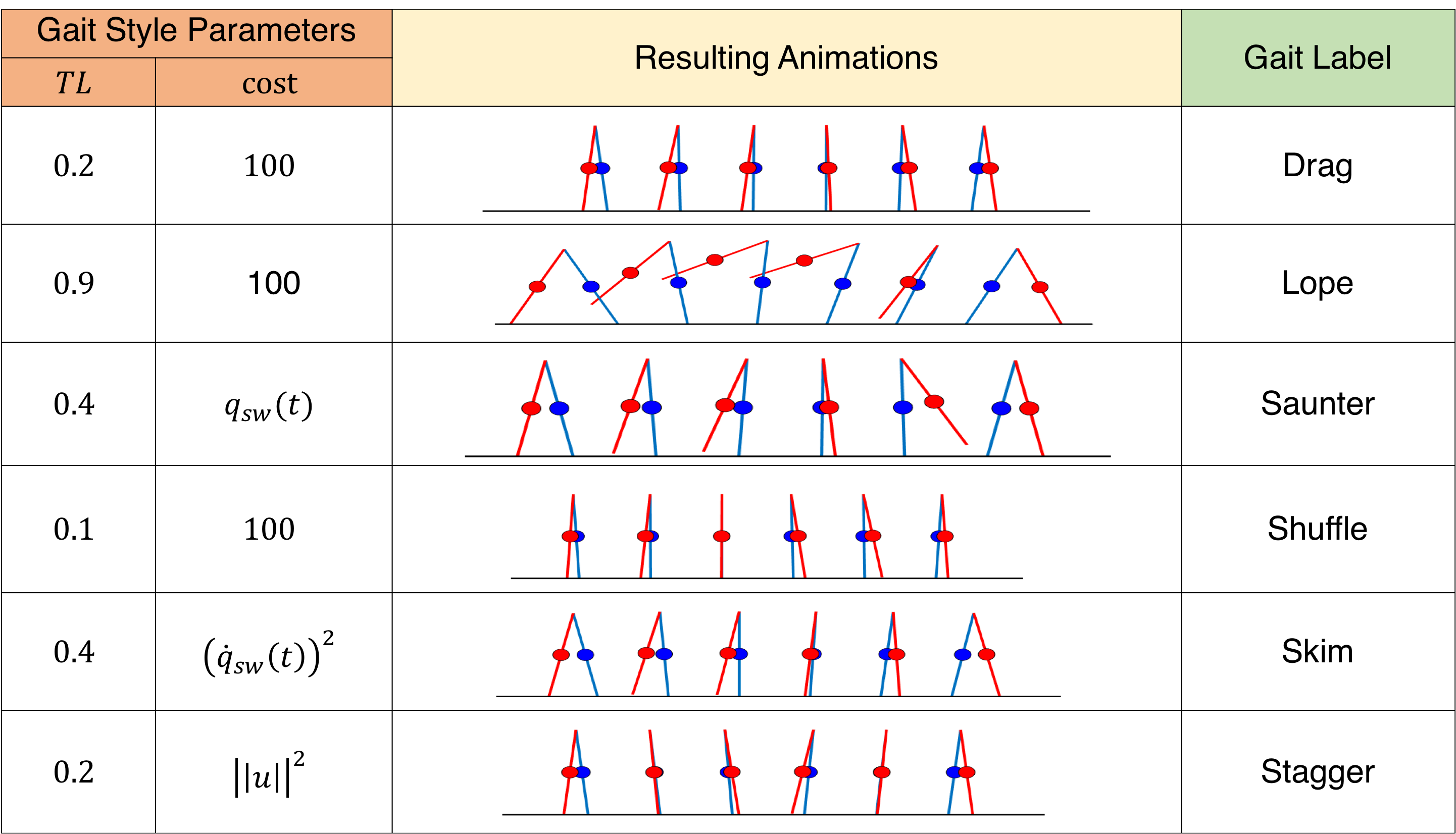}
    \caption{The set of labeled gaits, along with the corresponding optimization parameters, used for running the user study. The gaits are shown by static snapshots from the gait animation, overlayed throughout a single gait cycle.}
    \label{fig:gaits_surv}
\end{figure*}     

Next, trial user studies were run for validation of these gait labels. This involved training of the viewers for the meaning and sentence use of the words for the gait labels (the details of this process are described in the next section). The feedback and results from these trial studies were used to further refine the selection of gaits and their gait labels. After a couple of iterations, six gait styles with labels were selected for final validation with a bigger participant pool. The gaits chosen for the study are provided in Fig. \ref{fig:gaits_surv}. The words used for the gait labels, along with their meanings, and their use in sentences, are provided in Table \ref{fig:train_words}.

\begin{table*}[]
\sffamily 
\centering
\begin{tabular}{|c|l|l|}
\hline
\rowcolor[HTML]{C6E0B4} 
\multicolumn{1}{|c|}{Gait Label } & \multicolumn{1}{c|}{Meaning}                                                                                                                             & \multicolumn{1}{c|}{ \cellcolor[HTML]{C6E0B4} Training Sentence}                                                                                                          \\\cellcolor[HTML]{C6E0B4} &\cellcolor[HTML]{C6E0B4} &\cellcolor[HTML]{C6E0B4} \\ \hline
Drag       & \begin{tabular}[l]{@{}l@{}}To trail, to hang with its weight, while moving or being moved;\\ to move with friction on the ground or surface. \end{tabular}  & I have to drag myself out of bed each day. \cite{oxford_dict_eng2010}                                                                                                                                    \\ \hline
Lope       & To run or move with a long bounding stride.                                                                                          & \begin{tabular}[l]{@{}l@{}}She put the horse into a lope and headed for the\\ shed. \cite{lope_example} \end{tabular}                                                                                \\ \hline
Saunter    & \begin{tabular}[l]{@{}l@{}}Walk leisurely and with no apparent aim. \end{tabular}                                                   & \begin{tabular}[l]{@{}l@{}}In June, some flights were delayed at Kennedy\\ when about 100 turtles, seeking a place to lay\\ their eggs, sauntered across a runway. \cite{saunter_example} \end{tabular} \\ \hline

Shuffle    & \begin{tabular}[l]{@{}l@{}}Walk by dragging one’s feet along or without lifting them fully\\ from the ground. \end{tabular}           & \begin{tabular}[l]{@{}l@{}}I stepped into my skis and shuffled to the edge of\\ the steep slope. \cite{shuffle_example} \end{tabular}                                                                   \\ \hline
Skim       & \begin{tabular}[l]{@{}l@{}}To move, glide, fly or float, lightly and rapidly over or along (the\\ ground, etc.)\end{tabular}        & \begin{tabular}[l]{@{}l@{}}The swallows skimmed along the surface of\\ water. (Modified from \cite{kellogg1892john})\end{tabular}                                                                                      \\ \hline
Stagger    & \begin{tabular}[l]{@{}l@{}}To sway involuntarily from side to side when trying to stand or\\ walk erect. \end{tabular}                & \begin{tabular}[l]{@{}l@{}}A young woman staggered towards the\\ landlady, and then fell down in a swoon. (Modified from \cite{whitnell1982ring})\end{tabular}                                                        \\ \hline
\end{tabular}
\caption{The participants of the user studies were provided with the meanings of the gait labels and their uses in sentences during training and during the rating portion of the study.}
\vspace{-.1in}
    \label{fig:train_words}
\end{table*}

In the 6 gaits selected for user validation, the label of ``Drag'' was chosen because of a slow movement and a minimal lift of swing leg from ground. Similarly, ``Lope'' was given because the swing leg takes a relatively bigger step length as if trying a long stride. The label of ``Saunter'' was chosen for its respective gait because the leg movement in this appeared leisurely. Small steps and minimal lift from ground signified ``Shuffle''. ``Skim'' was chosen for its gait because of a seemingly gliding movement profile. ``Stagger'' was given for its gait because it represented slow walking with multiple swings of the swing leg before taking the step.  

\section{Validation of Gait Styles with Lay Viewers}
\label{sec_validate}
For final validation from human viewers, we prepared a user study with a bigger participant pool. In designing the study, which was used for all trial runs as well, one challenge was to take the attention away from the morphology of the biped and focus on the pattern of the movement profile. This was done through a training period. 

Just as a trained ballet dancer is able to distinguish more features of a ballet than a novice, we knew that users freshly exposed to the gait styles would need some time to resolve pattern in the movement.  This is what the training period provided.  First, users were given a tutorial on the human analogy to the planar platform, as in Fig. \ref{fig:knobs+vals}.  Users needed to pass a two question quiz (asking, for example, ``How many `legs' does the robot have?'') before advancing (via as many tries as it took).  Then, users were shown each gait and its corresponding label, along with a definition and example use of each label (provided in Table \ref{fig:train_words}).
The details in Table \ref{fig:train_words} were also provided as a referral document, via an on screen link, for the participants throughout the user study. 

After training, the rest of the questions asked were about the gait animation labeling. For each of the 6 gait animations, each participant was asked to rate its gait label on a scale of 1 to 7 (where 1 corresponded to the least accurate and 7 corresponded to the most accurate label). The participants were also required to provide an explanation for their rating as a typed description. Each gait animation was asked about on a separate page. The sequence of these gait animations was randomized for each of the participants. 

After questions about each animation, a human verification question was asked to ensure that the participants were reading through the questions carefully. Responses with wrong answers to any of these questions were invalidated. Finally, there were 9 standard demographic questions about the participant's background. 

\begin{figure}
     \centering        
	     \includegraphics[width=1\columnwidth]{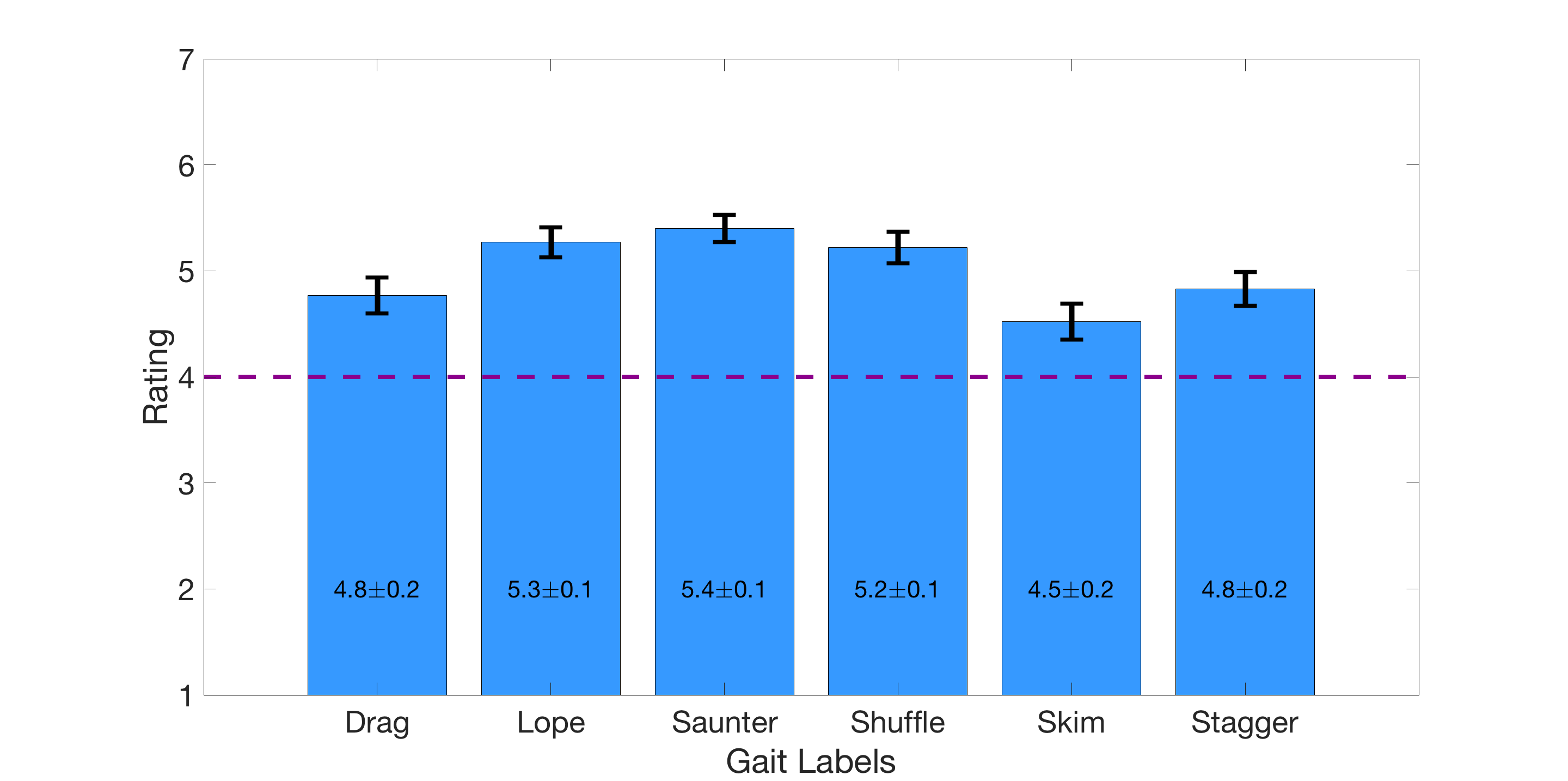}     
     \caption{Results of gait validation.  The average rating score from the user study for each gait label is plotted. A rating of 1 indicates poor fit between gait and label, and a rating of 7 shows the most accurate fit. Error bars indicate one standard deviation in the scores. Since all scores are over the middle rating of 4, it shows that the participants, on average, agreed with the gait labels.}
     \label{score}
     \vspace{-0.2in}
\end{figure} 

The study was implemented as a questionnaire in Qualtrics \cite{qualtrics} containing a total of 29 questions. 
A total of 100 qualifying participants were recruited using Amazon Mechanical Turk (MTurk) \cite{mturk}. The pool of user study participants was restricted to workers who had previously completed at least 100 HITs with an approval rating of at least 90\%. Only those workers were recruited who were not part of the trial studies discussed in Section \ref{sec_synthesis}. The participant pool for this user study comprised of 48 females and 52 males ranging between the ages of 21 and 68. About 71\% of the participants were native English speakers and others were fluent in English too.  
   
The results of the user study are given in Fig. \ref{score}. All the labels got an above average score of over 4, indicating overall agreement with each label. This indicates that the framework produced meaningful behaviors. The label ``Saunter'' received the highest of all. The reason for this may be linked to the unique profile of the swing leg moving past the landing position and coming back to it. The lowest score was received by ``Skim''. This may be because the small amplitude of swing leg motion is similar to that in ``Shuffle'' and ``Drag'', creating less distinction between these gaits. 
   
Some participants had specific comments about the gaits as well. For example, for ``Stagger'' one of the participants giving low rating commented, ``you can't sway sided to side in 2d''. For Drag, ``This just looked like it was in slow motion.  It was difficult to say why it was moving slow although dragging is one possibility...'', was one comment from a low commenter. On the other hand, some commenters appreciated the features in the movement that related to the gait labels. One commenter in favor of ``Stagger'' mentioned, ``Though the stride is short, the figure did seem like they were about to topple over.'', pointing out the instability in the movement. Another commenter giving higher score to ``Drag'' said, ``It did have a slow tired pace and the feet never leave the ground as if they are too heavy to lift''. 

In light of these comments and the overall rating scores, it can be concluded that while the model is impoverished and can not emulate the human movement originally ascribed by the labels exactly, majority of the participants agreed to these labels for their respective gait animations. Therefore, it can be concluded that our framework is capable of generating expressive gaits which can be linked to human walking behaviors.          

\section{Towards Expressive Bipedal Robots} \label{conc}
 In this paper, we have presented an under-actuated planar model of a bipedal robot, using a standard compass-like planar biped model, and a gait generation framework that allows for many variable gaits to be generated. Using this framework and a novel application of embodied movement analysis, a set of gaits has been identified to have meaningful similarity to human walking behavior. A user study has been conducted to validate these gait labels, all of which receive ratings of agreement with the labels by lay viewers as well as experts.  This indicates that this framework is capable of generating gaits that are meaningful to human viewers.
	
The overall goal of this work is to develop a framework for expressive gait generation. Variable movements are likely necessary to have a desired effect on human viewers and imitate social practices in human-facing settings. This behavior is important to social interactions and may facilitate, alongside other efforts, better acceptance of and interaction with walking robots.

\begin{figure}
	\centering
    \includegraphics[width = \columnwidth]{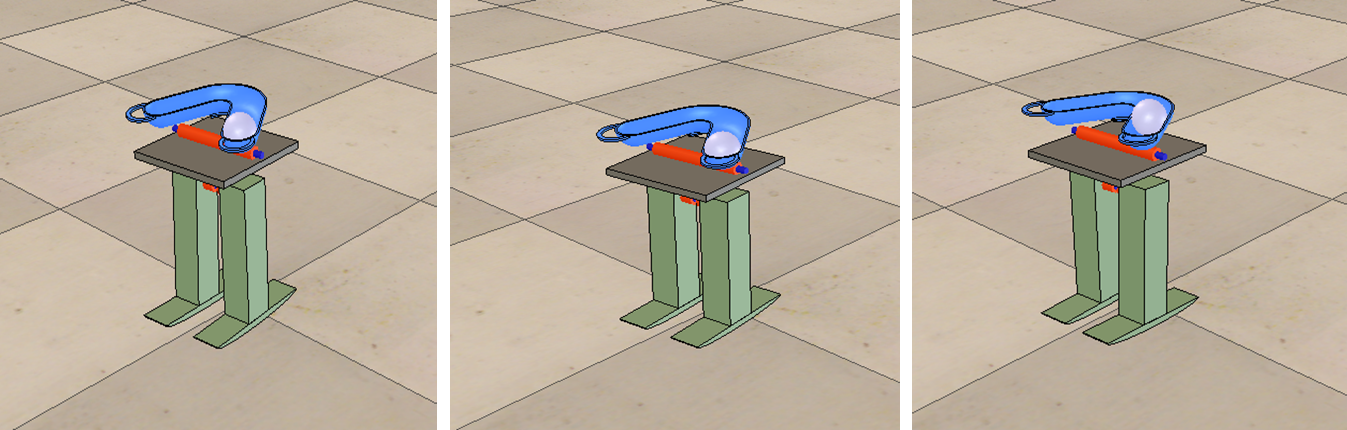}
    \caption{A step of walk in simulation on a prototype bipedal robot under development in conjunction with this work described in \cite{umer2016lbms}.}
    \label{fig_3Dbiped}
\end{figure}

Ideas in embodied movement analysis like Bartenieff Basic Six inspire developing bipedal robots with core-located actuation \cite{umer2016lbms} (Fig. \ref{fig_3Dbiped}) and in future, we are working towards making a hardware prototype. Furthermore, we would like to investigate the expressivity of that design following a similar process of assigning gait labels and conducting user studies. Another line of interest will be investigating functional advantages of such a design by comparing the range of walking styles admissible in them in comparison with simpler designs.

Other possible directions include adding more high-level movement ideas from LBMS in our framework, like Effort. This way, the gait styles already generated in our framework can have further variations because Effort parameters can add dynamic quality, clarifying further the movement's expression. Moreover, the gait styles produced here may be able to generate affective responses in humans, particularly when situated in corresponding contexts as in \cite{heimerdinger2017influence}.  Finally, selecting an appropriate gait from a large palette is a research challenge that needs to be addressed.

Humans make judgments about their counterparts based on their movement profile and actions. Therefore, when introducing robots in human-facing, social scenarios, it is important to develop the expressive capability of robots by equipping them with a wide range of movement options. With this capability, robots may have more meaningful interactions with human colleagues, engendering communication, acceptance, and trust.

	\appendix{} \label{matrices}
		The modeling matrices of Eq. \ref {swing_model} are as follows:
		\begin{align} 
		D_s =
		\begin{bmatrix}
		(\frac{5}{4}m)r^2  & - \frac{m}{2}r^2c_{12} \\
		- \frac{m}{2}r^2c_{12} &         \frac{m}{4}r^2 \\		
		\end{bmatrix},
		\end{align}
		where $c_{12}=\cos(q_{st}-q_{sw})$ and
		
		\begin{align}
		C_s=
		\begin{bmatrix}
		0 & -\frac{m}{2}r^2\dot q_{sw}s_{12}  \\
		\frac{m}{2}r^2\dot q_{st} s_{12} &  0 \\
		\end{bmatrix},
		\end{align}
		where $s_{12} = \sin(q_{st}-q_{sw}))$ and
		\begin{align}
		G_s=
		\begin{bmatrix}
		-(\frac{3m}{2})gr\sin(q_{st}) \\
		\frac{m}{2}gr \sin(q_{sw}))\\
		\end{bmatrix},
		\end{align}

		The matrices from Eq. \ref{impact_model} are as follows:
		\begin{align}
		D_e=
		\begin{bmatrix}
		D_s & D_{12} \\
		D^T_{12} & D_{22} \\
		\end{bmatrix}
		\end{align}
		\begin{align}
		D_{12}=
		\begin{bmatrix}
		\frac{3m}{2}r\cos q_{st} & -\frac{3m}{2}r\sin q_{st} \\
		-\frac{m}{2}r\cos q_{sw} & \frac{mr}{2}\sin q_{sw} \\
		M_t & 0
		\end{bmatrix}
		\end{align} and 
		\begin{align}
		D_{22}=
		\begin{bmatrix}
		2m & 0\\
		0 & 2m
		\end{bmatrix}.        
		\end{align}
	Similarly,        
		\begin{align}
		C_e=
		\begin{bmatrix}
			C_s & 0_3\\
			C_1 & 0_2
		\end{bmatrix},
		\end{align}
		where
		\begin{align}
		C_1=
		\begin{bmatrix}
		-\frac{3m}{2} r \dot q_{st} \sin q_{st} & \frac{mr}{2} \dot q_{sw} \sin q_{sw} \\
		-\frac{3m}{2} r \dot q_{st} \cos q_{st} & \frac{mr}{2} \dot q_{sw} \cos q_{sw} \\
		\end{bmatrix},
		\end{align}		
	and        
		\begin{align}
		G_e=
		\begin{bmatrix}
		G_s\\
		G_1\\			
		\end{bmatrix},
		\end{align} where\\		
		\begin{align}
		G_1=
		\begin{bmatrix}
		0\\
		2mg\\
		\end{bmatrix}.
		\end{align}

\subsection*{Acknowledgment}		
This work was conducted under IRB \#17697 and funded by National Science Foundation (NSF) grant \#1701295. The authors would like to thank Prof. Hae Won Park for useful discussions about the controller design and trajectory optimization and Prof. Joshua Schultz for useful discussions about how this control scheme might be implemented through a physical mechanism.

\subsection*{Compliance with Ethical Standards:}
Funding: This study was funded by the National Science Foundation (NSF) grant \#1701295.
Conflict of Interest: A. LaViers owns stock in AE Machines, an automation software company.

\bibliographystyle{IEEEtran}
\bibliography{papers}
\vspace{-0.5in}
\begin{IEEEbiography}[{\includegraphics[width=1in,height=1.25in,clip,keepaspectratio]{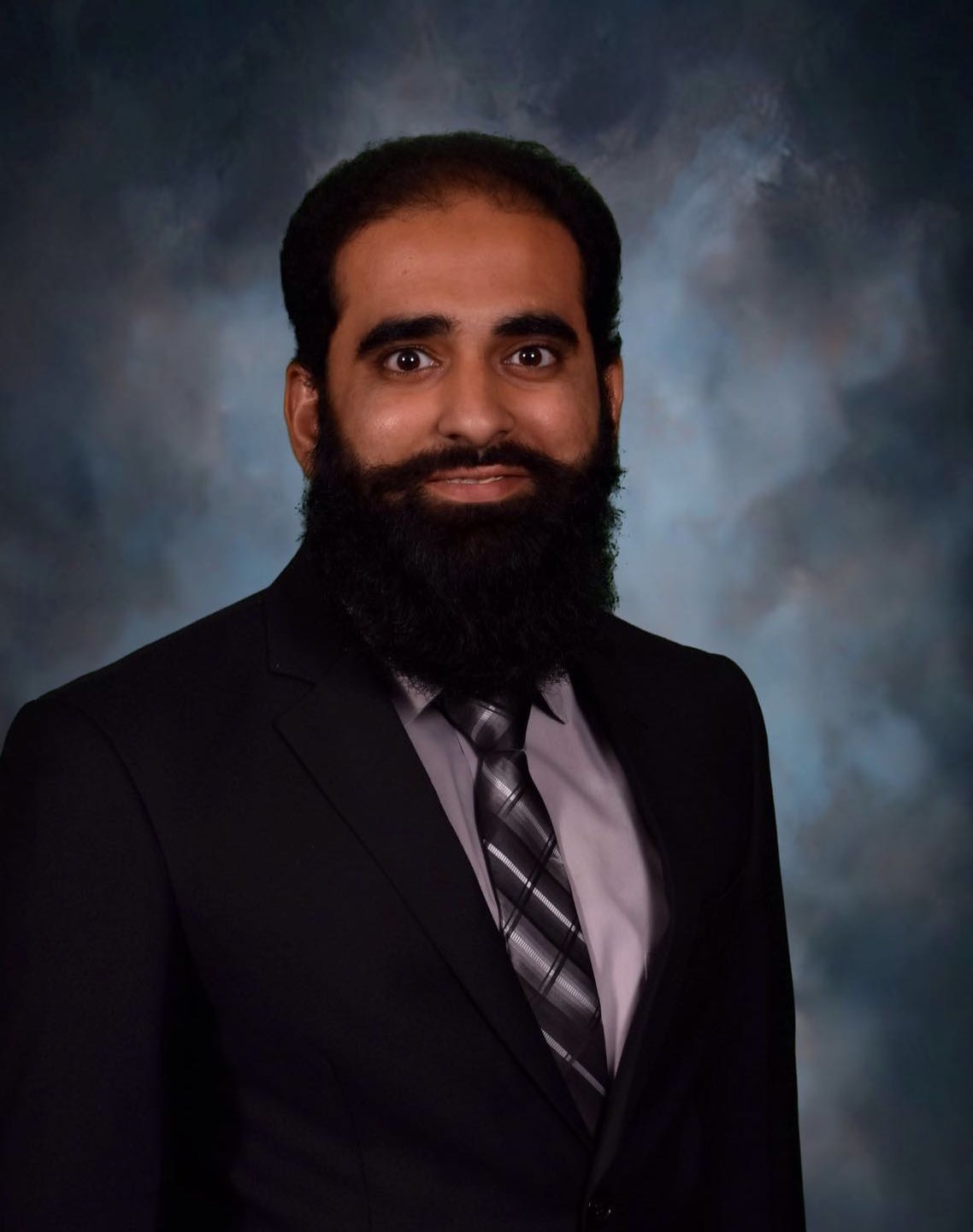}}]{Umer Huzaifa} is a PhD candidate in Mechanical Engineering at the University of Illinois at Urbana-Champaign. He received his MS from University of Illinois in 2016 and BS from National University of Sciences and Technology, Pakistan in 2012. He was one of the finalists for the Best Paper Award in International Conference on Biomedical Robotics and Biomechatronics (BioRob), 2016. His research interests are expressive robotics, and dynamics and control in legged locomotion. 
\end{IEEEbiography}
\begin{IEEEbiography}
[{\includegraphics[width=1in,height=1.25in,clip,keepaspectratio]{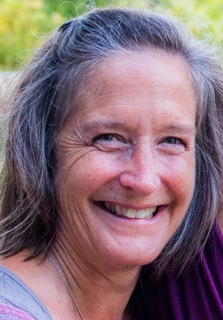}}]{Catherine (Cat) Maguire} is a movement educator, dance artist, Certified Movement Analyst and Registered Somatic Movement Educator, teaching in the Charlottesville area and at the University of Maryland and internationally for the Laban/Bartenieff Institute of Movement Studies (LIMS).  After graduating Phi Beta Kappa from Wesleyan University with honors in dance and psychology, Maguire earned a Certified Movement Analyst (CMA) degree with LIMS in New York City. For the next eight years, she was the artistic director of Offspring Dance Company in New York City and the founder and head of the dance program at Drew University in Madison, NJ. While executive director of Piedmont Council of the Arts in Charlottesville, Maguire continued to teach, choreograph and perform throughout Central Virginia and internationally. From 2003 to 2010, Maguire was assistant professor of dance at Piedmont Virginia Community College (PVCC), where she was an integral part of the development and implementation of the associate's degree in dance, the only one of its kind in the Virginia Community College System. Currently Maguire teaches group classes and individual sessions designed to foster self-expression, body connectivity and transformation through movement.
\end{IEEEbiography}
\vspace{-4.5in}
\begin{IEEEbiography}
[{\includegraphics[width=1in,height=1.25in,clip,keepaspectratio]{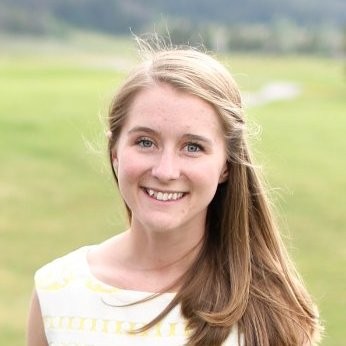}}]{Amy LaViers} is an assistant professor in the Mechanical Science and Engineering Department at the University of Illinois at Urbana-Champaign and director of the Robotics, Automation, and Dance (RAD) Lab. She is the recipient of a 2015 DARPA Young Faculty Award (YFA) and 2017 Product Design of the Year at the 4th Revolution Awards in Chicago for her start up AE Machines. She completed a Certification in Movement Analysis (CMA) in 2016 at the Laban/Bartenieff Institute of Movement Studies (LIMS). Prior to UIUC she held a position as an assistant professor for two years in systems and information engineering at the University of Virginia. She completed her Ph.D. in electrical and computer engineering at Georgia Tech. Her dissertation included a live performance exploring the concepts of style she developed there. Her research began at Princeton University where she earned a certificate in dance and a degree in mechanical and aerospace engineering and explores how choreographic and somatic practices can inform the development of more expressive robotic systems.  
\end{IEEEbiography}
\end{document}